\documentclass[11pt]{article}

\usepackage[]{acl}

\usepackage{times}
\usepackage{latexsym}

\usepackage{microtype}
\usepackage{graphicx}
\usepackage{adjustbox}
\usepackage{multirow}
\usepackage{paralist}
\usepackage[inline]{enumitem}
\usepackage{graphicx}
\usepackage{url}
\usepackage{textcomp}
\usepackage{amssymb}
\usepackage{caption}
\usepackage{subcaption}
\usepackage{color,soul}
\usepackage{paralist}
\usepackage{float}
\usepackage{placeins}
\usepackage{aliascnt}
\usepackage{mathtools}
\usepackage{amsmath}
\usepackage{hyperref}
\usepackage{fontawesome5}
\usepackage{framed}
\usepackage{booktabs}
\usepackage{graphicx}

\usepackage{algpseudocode}
\usepackage[ruled,vlined]{algorithm2e}
\usepackage[dvipsnames]{xcolor}
\usepackage[T1]{fontenc}

\usepackage{rotating}

\usepackage[utf8]{inputenc}

\usepackage{array}
\newcolumntype{C}[1]{>{\centering\arraybackslash}p{#1}}

\usepackage{microtype}

\usepackage{inconsolata}

\usepackage{graphicx}

\title{\includegraphics[scale=0.02]{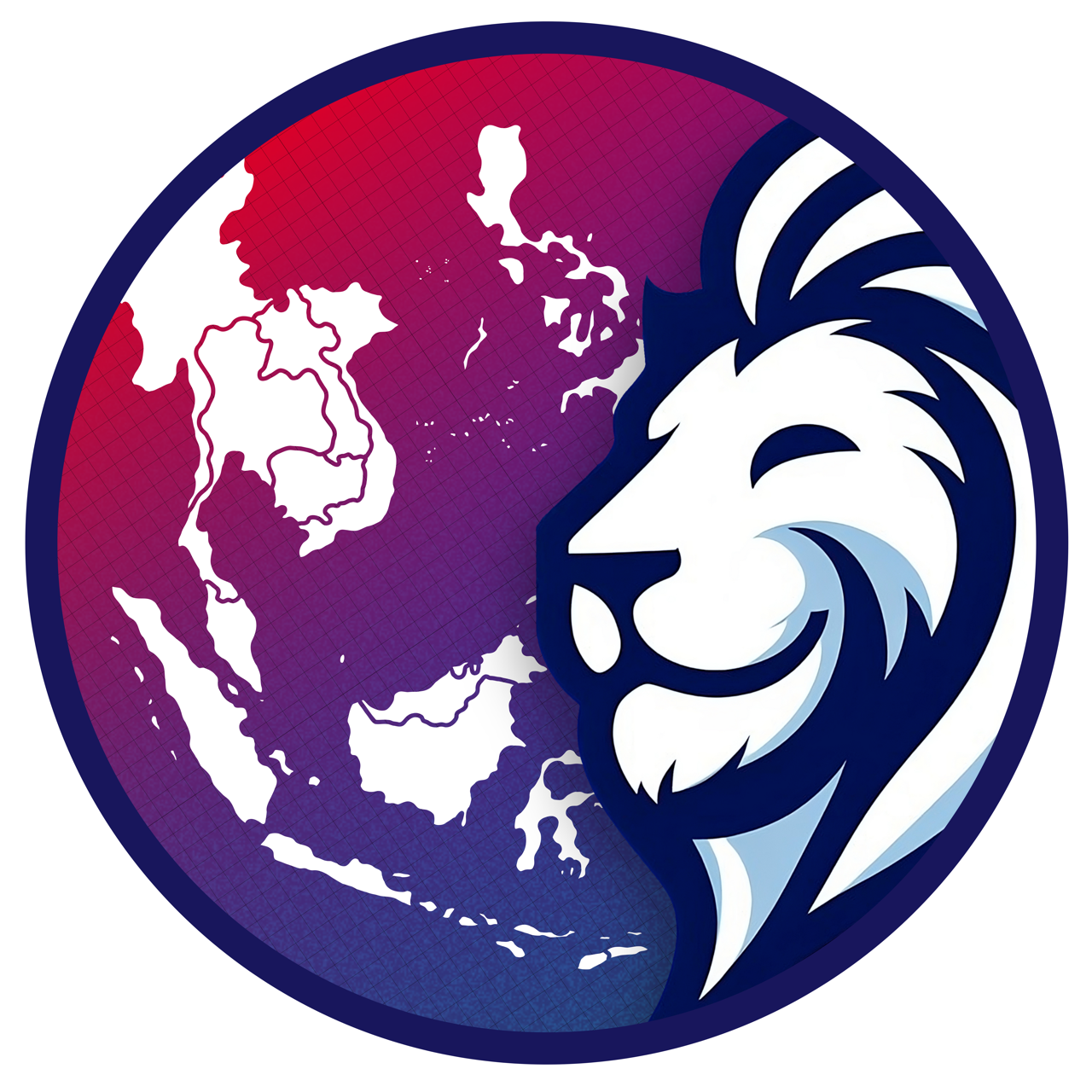}SEA-LION-Embedding: Open and Reproducible Text Embeddings for Southeast Asia}

\author{
 \textbf{Peerat Limkonchotiwat\textsuperscript{1}},
 \textbf{Raymond Ng\textsuperscript{1}, Sarana Nutanong\textsuperscript{2}, Jian Gang Ngui \textsuperscript{1}}
\\
 \textsuperscript{1}AI Singapore
 \textsuperscript{2}School of Information Science and Technology, VISTEC
 \\
 \texttt{peerat@aisingapore.org}
 \\
 \href{https://github.com/aisingapore/SEA-LION-Embedding}{\faGithub\ GitHub}
\quad
\href{https://huggingface.co/collections/aisingapore/sea-lion-modernbert-and-embedding}{\includegraphics[height=1em]{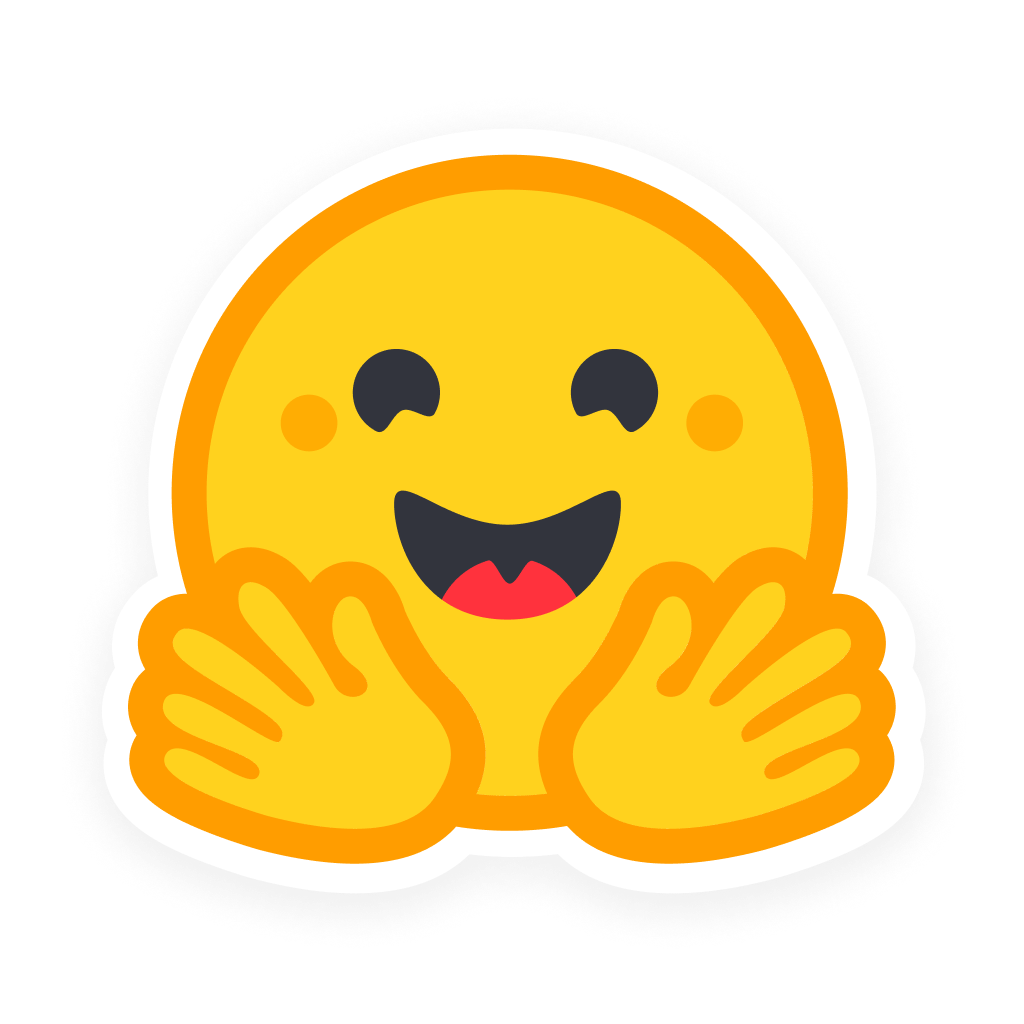}\ Hugging Face}
}

\begin{document}
\maketitle
\begin{abstract}
Text embeddings are fundamental to many downstream applications, making robustness important for real-world NLP.
However, most recent state-of-the-art embedding models are not reproducible because they rely on closed or undisclosed training data, and they remain insufficiently robust for Southeast Asian languages.
We present SEA-LION-Embedding, a fully open and reproducible text-embedding pipeline for Southeast Asian languages trained only on publicly available data, and use it to study three core factors of robust embedding design: data composition, training objective, and base encoder initialization.
SEA-LION-Embedding achieves state-of-the-art results on SEA-BED while enabling systematic and reproducible analysis of robust text embeddings for the region.

\end{abstract}

\section{Introduction}

%
Text embeddings aim to build a semantic space that preserves meaning across tasks and languages.
For Southeast Asian (SEA) languages, linguistic diversity is high, and gains from high-resource settings may not transfer~\cite{singh-etal-2025-global,susanto-etal-2025-sea}.
Recent results from SEA-BED~\cite{ponwitayarat2025seabedsoutheastasiaembedding} show that models that perform strongly on global benchmarks such as MMTEB~\cite{enevoldsen2025mmteb} still lag in SEA evaluations, motivating a focused study of robust, reproducible embeddings for the region.

Prior work improves multilingual text embeddings by (i) scaling or diversifying training data~\cite{wang-etal-2024-improving-text, wang2024multilinguale5textembeddings, zhang2025qwen3embeddingadvancingtext,hu2025kalmembedding,jinav5}, (ii) strengthening contrastive/distillation objectives~\cite{chen-etal-2024-m3, wang2024multilinguale5textembeddings, hu2025kalmembedding, zhao2026kalmembeddingv}, and (iii) initializing from stronger pretrained encoders~\cite{wang2024textembeddingsweaklysupervisedcontrastive,wang2024multilinguale5textembeddings}.
Together, these map to three core questions: what the model learns from, how it learns, and where it starts.

In this paper, we introduce SEA-LION-Embedding: a transparent, reproducible text-embedding pipeline for Southeast Asian languages trained solely on public data (245M text pairs; 14M instruction texts).
Table~\ref{tab:overview} contrasts prior high-performing multilingual embedding models with our open pipeline, enabling systematic, controlled comparisons.
This transparency matters in practice: open \emph{code} exposes objectives and training details, open \emph{data} supports re-training and ablations, and a reproducible \emph{environment} makes comparisons and failure analysis reliable.
\begin{table}[htbp]
\centering
\vspace{-1mm}
\fontsize{7pt}{13pt}
\selectfont
\setlength{\tabcolsep}{2.2pt}
\renewcommand{\arraystretch}{0.9}
\scalebox{.98}{
\begin{tabular}{l C{0.8cm} C{0.8cm} C{0.9cm} | C{1.0cm}}
\hline
\textbf{Model}
& \shortstack[c]{\textbf{Open}\\[-2pt]\textbf{Code}}
& \shortstack[c]{\textbf{Open}\\[-2pt]\textbf{Data}}
& \shortstack[c]{\textbf{Reprod.}\\[-2pt]\textbf{Env.}}
& \shortstack[c]{\textbf{SEA-BED}\\[-2pt]\textbf{Lang-Avg}} \\ \hline
Qwen3-Embedding                   & No               & No                  & No                  & 0.607           \\
BGE-M3                            & Partial               & \textbf{Full}               & \textbf{Full}               & 0.765           \\ 
$m$E5-Large-Instruction & No               & Partial             & \textbf{Full}                  & 0.789           \\ 
Jina-Embedding-v5                 & No               & No             & No             & 0.694           \\ 
harrier-oss-v1                 & No               & No             & No             & 0.781           \\ 
SEA-LION-Embedding (Ours)              & \textbf{Full}                 & \textbf{Full}               & \textbf{Full}               & \textbf{0.800}           \\ \hline
\end{tabular}}
\vspace{-1mm}
\caption{Comparison with prior multilingual text embeddings (all models are 0.6B parameters). SEA-LION-Embedding achieves the best SEA-BED language-average score while also providing open code, open data, and a fully reproducible experimental environment.}
\label{tab:overview}
\vspace{-3mm}
\end{table}

Using this pipeline, \emph{we study robustness} as performance consistency across languages and task types on benchmarks.
We analyze three design factors: data composition, objective design, and base-encoder initialization under different pretraining conditions. 
Our model achieves state-of-the-art results on the held-out SEA-BED benchmark with fully reproducible experiments.
%




Our contributions are \begin{inparaenum}[(i)]
    \item \textbf{Reproducible framework}: SEA-LION-Embedding is a public-data-only pipeline with released artifacts for re-training and controlled comparisons.
    \item \textbf{Empirical insights}: We isolate the effects of data composition, objective design, and base-encoder initialization on SEA robustness.
    \item \textbf{Robust performance}: We achieve state-of-the-art SEA-BED results and distill a reusable design recipe.
\end{inparaenum}

\section{Proposed SEA-LION-Embedding Pipeline}

\begin{figure}[htbp]
  \centering
  \vspace{-2mm}
  \includegraphics[width=0.9\linewidth]{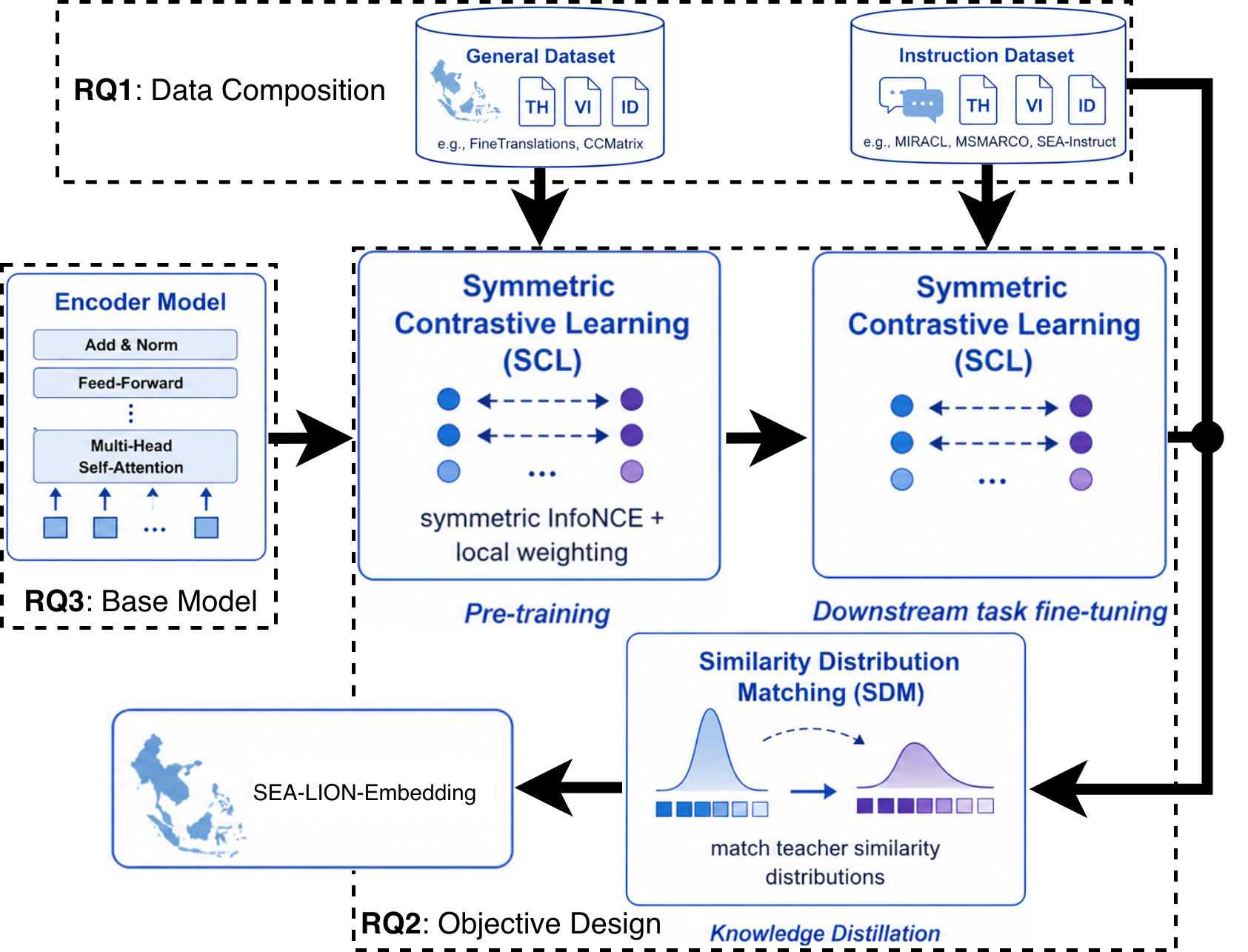}
  \vspace{-3mm}
  \caption{The overview of SEA-LION-Embedding}
  \vspace{-7mm}
  \label{fig:overview}
\end{figure}

Figure~\ref{fig:overview} presents our training pipeline for SEA-LION-Embedding.
This structure serves as a conceptual framework for systematically examining three critical components of robust text embeddings: \textbf{RQ1:} Data Composition, \textbf{RQ2:} Objective Design, and \textbf{RQ3:} Base Model.

\subsection{RQ1: Data Composition}

We propose that robust SEA text embeddings require training data with both broad regional coverage and task-oriented supervision.
Broad coverage helps a shared semantic space capture linguistic variation across Southeast Asian languages, while task-oriented supervision supports diverse downstream applications.
We therefore use two data categories: \textbf{general datasets} (245M samples) for SEA language coverage and \textbf{instruction datasets} (14M samples) for task awareness and generalization.
For \textbf{general datasets}, we use SEA-language resources such as FineTranslations~\cite{penedo2026finetranslations}, CCMatrix~\cite{schwenk-etal-2021-ccmatrix}, etc.
For \textbf{instruction datasets}, we use multilingual downstream datasets such as MIRACL~\cite{zhang-etal-2023-miracl}, \href{https://huggingface.co/datasets/aisingapore/SEA-Instruct-2602}{SEA-Instruct}, etc., prepending task-specific instructions to improve task awareness and generalization.
Both data types are formatted as triplets (anchor, positive, negative) or text pairs (anchor, positive), following the original dataset format.

\begin{table*}[t!]
\centering
\vspace{-3mm}
\fontsize{7pt}{13pt}
\selectfont
\scalebox{0.9}{
\begin{tabular}{lrrrrrrrrrrr}
\hline
\multicolumn{1}{l|}{\textbf{Model}} &
  \multicolumn{1}{c|}{\textbf{Ind}} &
  \multicolumn{1}{c|}{\textbf{Tha}} &
  \multicolumn{1}{c|}{\textbf{Vie}} &
  \multicolumn{1}{c|}{\textbf{Fil}} &
  \multicolumn{1}{c|}{\textbf{Mya}} &
  \multicolumn{1}{c|}{\textbf{Tam}} &
  \multicolumn{1}{c|}{\textbf{Khm}} &
  \multicolumn{1}{c|}{\textbf{Zsm}} &
  \multicolumn{1}{c|}{\textbf{Lao}} &
  \multicolumn{1}{c|}{\textbf{Tet}} &
  \multicolumn{1}{c}{\textbf{Avg.}} \\ \hline \hline

\multicolumn{12}{l}{\textit{$\leq$300M parameters}} \\ \hline
\multicolumn{1}{l|}{embeddinggemma-300m} &
  \multicolumn{1}{r|}{\textbf{0.802}} &
  \multicolumn{1}{r|}{\textbf{0.806}} &
  \multicolumn{1}{r|}{\textbf{0.764}} &
  \multicolumn{1}{r|}{0.704} &
  \multicolumn{1}{r|}{0.571} &
  \multicolumn{1}{r|}{0.761} &
  \multicolumn{1}{r|}{0.641} &
  \multicolumn{1}{r|}{0.792} &
  \multicolumn{1}{r|}{0.530} &
  \multicolumn{1}{r|}{0.673} &
  0.704 \\ 
\multicolumn{1}{l|}{harrier-oss-v1-270m} &
  \multicolumn{1}{r|}{0.779} &
  \multicolumn{1}{r|}{0.765} &
  \multicolumn{1}{r|}{0.745} &
  \multicolumn{1}{r|}{\textbf{0.774}} &
  \multicolumn{1}{r|}{0.724} &
  \multicolumn{1}{r|}{0.745} &
  \multicolumn{1}{r|}{0.739} &
  \multicolumn{1}{r|}{0.830} &
  \multicolumn{1}{r|}{0.753} &
  \multicolumn{1}{r|}{0.\textbf{679}} &
  0.753 \\ 
\multicolumn{1}{l|}{SEA-LION-Embedding-ModernBERT-300M (Ours)} &
  \multicolumn{1}{r|}{0.787} &
  \multicolumn{1}{r|}{0.789} &
  \multicolumn{1}{r|}{0.761} &
  \multicolumn{1}{r|}{\textbf{0.774}} &
  \multicolumn{1}{r|}{\textbf{0.739}} &
  \multicolumn{1}{r|}{\textbf{0.774}} &
  \multicolumn{1}{r|}{\textbf{0.773}} &
  \multicolumn{1}{r|}{\textbf{0.838}} &
  \multicolumn{1}{r|}{\textbf{0.802}} &
  \multicolumn{1}{r|}{0.563} &
  \textbf{0.760$^\dagger$} \\ \hline
\multicolumn{12}{l}{\textit{$>$300M parameters}} \\ \hline
\multicolumn{1}{l|}{multilingual-e5-large-instruct} &
  \multicolumn{1}{r|}{0.795} &
  \multicolumn{1}{r|}{0.811} &
  \multicolumn{1}{r|}{0.780} &
  \multicolumn{1}{r|}{0.784} &
  \multicolumn{1}{r|}{\textbf{0.792}} &
  \multicolumn{1}{r|}{0.771} &
  \multicolumn{1}{r|}{0.781} &
  \multicolumn{1}{r|}{0.846} &
  \multicolumn{1}{r|}{0.839} &
  \multicolumn{1}{r|}{0.694} &
  0.789 \\ 
\multicolumn{1}{l|}{Qwen3-Embedding-0.6B} &
  \multicolumn{1}{r|}{0.756} &
  \multicolumn{1}{r|}{0.759} &
  \multicolumn{1}{r|}{0.751} &
  \multicolumn{1}{r|}{0.491} &
  \multicolumn{1}{r|}{0.631} &
  \multicolumn{1}{r|}{0.610} &
  \multicolumn{1}{r|}{0.441} &
  \multicolumn{1}{r|}{0.695} &
  \multicolumn{1}{r|}{0.298} &
  \multicolumn{1}{r|}{0.634} &
  0.607 \\ 
\multicolumn{1}{l|}{Qwen3-Embedding-8B} &
  \multicolumn{1}{r|}{0.797} &
  \multicolumn{1}{r|}{\textbf{0.815}} &
  \multicolumn{1}{r|}{0.790} &
  \multicolumn{1}{r|}{0.749} &
  \multicolumn{1}{r|}{0.781} &
  \multicolumn{1}{r|}{0.760} &
  \multicolumn{1}{r|}{0.755} &
  \multicolumn{1}{r|}{0.824} &
  \multicolumn{1}{r|}{0.782} &
  \multicolumn{1}{r|}{0.674} &
  0.773 \\ 
\multicolumn{1}{l|}{bge-m3 (Dense)} &
  \multicolumn{1}{r|}{0.781} &
  \multicolumn{1}{r|}{0.776} &
  \multicolumn{1}{r|}{0.759} &
  \multicolumn{1}{r|}{0.731} &
  \multicolumn{1}{r|}{0.758} &
  \multicolumn{1}{r|}{0.775} &
  \multicolumn{1}{r|}{0.762} &
  \multicolumn{1}{r|}{0.825} &
  \multicolumn{1}{r|}{0.823} &
  \multicolumn{1}{r|}{0.655} &
  0.765 \\
\multicolumn{1}{l|}{Cohere-embed-multilingual-v3.0} &
  \multicolumn{1}{r|}{0.797} &
  \multicolumn{1}{r|}{0.810} &
  \multicolumn{1}{r|}{0.789} &
  \multicolumn{1}{r|}{0.761} &
  \multicolumn{1}{r|}{0.790} &
  \multicolumn{1}{r|}{0.789} &
  \multicolumn{1}{r|}{0.770} &
  \multicolumn{1}{r|}{0.824} &
  \multicolumn{1}{r|}{0.833} &
  \multicolumn{1}{r|}{0.668} &
  0.783 \\ 
\multicolumn{1}{l|}{jina-embeddings-v5} &
  \multicolumn{1}{r|}{0.764} &
  \multicolumn{1}{r|}{0.779} &
  \multicolumn{1}{r|}{0.761} &
  \multicolumn{1}{r|}{0.619} &
  \multicolumn{1}{r|}{0.726} &
  \multicolumn{1}{r|}{0.742} &
  \multicolumn{1}{r|}{0.615} &
  \multicolumn{1}{r|}{0.784} &
  \multicolumn{1}{r|}{0.474} &
  \multicolumn{1}{r|}{0.674} &
  0.694 \\ 
\multicolumn{1}{l|}{harrier-oss-v1-0.6b} &
  \multicolumn{1}{r|}{\textbf{0.809}} &
  \multicolumn{1}{r|}{0.794} &
  \multicolumn{1}{r|}{0.783} &
  \multicolumn{1}{r|}{0.795} &
  \multicolumn{1}{r|}{0.742} &
  \multicolumn{1}{r|}{0.777} &
  \multicolumn{1}{r|}{0.759} &
  \multicolumn{1}{r|}{\textbf{0.859}} &
  \multicolumn{1}{r|}{0.769} &
  \multicolumn{1}{r|}{\textbf{0.722}} &
  0.781 \\ 
\multicolumn{1}{l|}{SEA-LION-Embedding-E5-Large-600M (Ours)} &
  \multicolumn{1}{r|}{0.808} &
  \multicolumn{1}{r|}{0.813} &
  \multicolumn{1}{r|}{\textbf{0.795}} &
  \multicolumn{1}{r|}{\textbf{0.808}} &
  \multicolumn{1}{r|}{0.785} &
  \multicolumn{1}{r|}{\textbf{0.806}} &
  \multicolumn{1}{r|}{\textbf{0.797}} &
  \multicolumn{1}{r|}{0.853} &
  \multicolumn{1}{r|}{\textbf{0.841}} &
  \multicolumn{1}{r|}{0.697} &
  \textbf{0.800$^\dagger$} \\ 
  \hline 
  \hline
\end{tabular}}
\vspace{-3mm}
\caption{Performance of all models on SEA-BED using language-average scores. 
$\dagger$ indicates significant improvement over the previous SOTA within each group using McNemar's test. 
Task-level results are reported in Appendix~\ref{sec:sea_bed_appendix}.}
\label{tab:sea_bed_lang}
\vspace{-5mm}
\end{table*}

\subsection{RQ2: Objective Design}

We propose that robust SEA text embeddings require both local discrimination and global consistency. 
%
%
Accordingly, we use a two-stage objective: Symmetric Contrastive Learning (SCL) and Similarity Distribution Matching (SDM).

\subsubsection{Symmetric Contrastive Learning (SCL)} \label{sec:scl}
SCL learns text embeddings that are robust across languages and downstream tasks.
Unlike prior works that rely on vanilla contrastive learning~\citep{chen-etal-2024-m3,wang2024multilinguale5textembeddings,hu2025kalmembedding,zhao2026kalmembeddingv}, we adopt symmetric InfoNCE~\citep{gunther2024jinaembeddings2} with focal reweighting.
We define $h^s(\cdot)$ as the encoder that maps an input text to a $d$-dimensional embedding using mean pooling over the last layer. 
For each training sample, we assume an anchor ($a$), a positive ($p$), and optionally a negative ($n$). 
Given a batch of $N$ triplets $\{(a_i,p_i,n_i)\}_{i=1}^{N}$, we obtain $\ell_2$-normalized embeddings $z_i^a$, $z_i^p$, and, when available, $z_i^n$ from $h^s(\cdot):\mathcal{X}\to\mathbb{R}^d$. 
$\mathcal{L}_{scl}$ is defined as follows:
\vspace{-2mm}
\begin{equation}
\vspace{-2mm}
\small
\begin{aligned}
\mathcal{L}_{scl}=-\frac{1}{2N}\sum_{i}^{N} w_i\Bigg[ &\log \frac{e^{s(z_i^a,z_i^p)}}{\sum_{j}^{N} e^{s(z_i^a,z_j^p)}+e^{s(z_i^a,z_i^n)}} \\
&+\log \frac{e^{s(z_i^p,z_i^a)}}{\sum_{j}^{N} e^{s(z_i^p,z_j^a)}}\Bigg],
\end{aligned}
\end{equation}
where $s(\mathbf{u},\mathbf{v})=\mathbf{u}^\top\mathbf{v}/\tau$, $\tau$ is a temperature parameter, and $w_i=(1-j_i)^\gamma$ is the focal weight~\citep{zhao2026kalmembeddingv}, with $j_i$ denoting the softmax probability assigned to the anchor-positive pair. 
When $\gamma=0$, the loss reduces to uniform weighting, while a larger $\gamma$ places more emphasis on harder examples. 
We train first on general texts, then continue on instruction texts with the same objective. 
Without explicit negatives, the loss reduces to the standard in-batch symmetric contrastive objective.
%

\subsubsection{Similarity Distribution Matching (SDM)} \label{sec:kd_loss}
We further improve SEA-LION-Embedding with Similarity Distribution Matching (SDM).
Unlike prior text-embedding distillations~\citep{jinav5,zhao2026kalmembeddingv}, SDM explicitly matches student and teacher similarity distributions over a large memory queue.
Let $h^s(\cdot)$ and $h^t(\cdot)$ denote the student and teacher encoders.\footnote{To match the teacher embedding size, we add a linear projection to the student output.} In addition, we use the model from Section~\ref{sec:scl} as the student model and ~\citet{wang2024multilinguale5textembeddings} as the teacher model. 
We maintain a first-in-first-out memory queue $\mathcal{P}$ of size 65,536. 
At each training step, we update $\mathcal{P}$ by encoding positive ($p$) and negative ($n$) samples with $h^t(\cdot)$, adding them to the queue, and removing the oldest entries. 
We then compute the similarity distribution between an anchor ($a$) and $\mathcal{P}$ as:
\vspace{-1mm}
\begin{equation}
\vspace{-1mm}
\small
    [D(h(a),\mathcal{P},\tau)]_k = \frac{e^{s(h(a),\,\mathbf{p}_k)/\tau}}{\sum_{\mathbf{p}_k\in\mathcal{P}} e^{s(h(a),\,\mathbf{p}_k)/\tau}}
\tag{2}
\end{equation}
where $\mathbf{p}_k\in\mathcal{P}$ denotes the $k$-th queue entry. We define the student and teacher similarity distributions as $D^s=D(h^s(a),\mathcal{P},\tau^s)$ and $D^t=D(h^t(a),\mathcal{P},\tau^t)$, with $\tau^s<\tau^t$ so that the student distribution is sharper, making it more challenging for the student model to match the distribution. We then let the student's distribution mimic the teacher's distribution using KL Divergence: $\mathcal{L}_{sdm}=\mathrm{KL}(D^t\|D^s)$.

\subsubsection{Comparison with Prior Works}

Prior text embeddings~\cite{wang2024multilinguale5textembeddings,hu2025kalmembedding,zhao2026kalmembeddingv} typically use standard contrastive learning with anchor-positive optimization and one-way in-batch negatives.
In contrast, $\mathcal{L}_\text{scl}$ uses symmetric querying and focal reweighting to better exploit batch samples and emphasize hard pairs.
$\mathcal{L}_\text{sdm}$ further matches soft similarity distributions over a 65k-sample memory queue $\mathcal{P}$, encouraging a more globally consistent embedding space than batch-level contrastive learning alone.
However, $\mathcal{L}_\text{sdm}$ requires a strong teacher encoder, as unreliable teacher similarities can destabilize training.
Thus, we use $\mathcal{L}_\text{sdm}$ only for knowledge distillation, rather than replacing $\mathcal{L}_\text{scl}$.

\subsection{RQ3: Base Models}
For RQ3, we test whether our pipeline transfers across starting encoders.
We instantiate it on 300M and 600M models, namely \href{https://huggingface.co/aisingapore/SEA-LION-ModernBERT-300M}{SEA-LION-ModernBERT-300M} and E5-Large~\cite{wang2024textembeddingsweaklysupervisedcontrastive}, to study efficiency--performance trade-offs.
Section~\ref{sec:exp} further evaluates XLMR-Large, mmBERT-base, and SEA-LION-ModernBERT-base to test whether gains generalize beyond specific initializations.
Dataset details, task-specific instructions, data leakage, and hyperparameters are provided in Appendices~\ref{sec:hyperparameters} and ~\ref{sec:datasets}.

\section{Experimental Setup and Results}
\label{sec:exp}

\noindent
\textbf{Benchmarks.}
We evaluate robustness on SEA-BED~\cite{ponwitayarat2025seabedsoutheastasiaembedding}, an out-of-domain benchmark for SEA-LION-Embedding covering 10 Southeast Asian languages,\footnote{Indonesian: Ind, Thai: Tha, Vietnamese: Vie, Filipino: Fil, Burmese: Mya, Tamil: Tam, Khmer: Khm, Malay: Zsm, Lao: Lao, Tetum: Tet.} 9 task types, and 169 datasets.
We report average language- and task-level scores.
English (MTEB) and Chinese (CMTEB) results are given in Appendix~\ref{sec:mteb_and_cmteb_results}.

\noindent
\textbf{Competitive Methods.}
We compare SEA-LION-Embedding against the top models on the SEA-BED leaderboard:\footnote{\url{https://leaderboard.sea-lion.ai/embedding/SEA}} multilingual-e5-large-instruct~\cite{wang2024multilinguale5textembeddings}, Qwen3-Embedding-0.6 and -8B~\cite{zhang2025qwen3embeddingadvancingtext}, Cohere embed-multilingual-v3.0, BGE-M3 (Dense)~\cite{chen-etal-2024-m3}, embeddinggemma-300m~\cite{embedding_gemma_2025}, \href{https://huggingface.co/microsoft/harrier-oss-v1-270m}{harrier-oss-v1-270M} and \href{https://huggingface.co/microsoft/harrier-oss-v1-0.6b}{0.6B}, and jina-embeddings-v5~\cite{jinav5}.

%
%

\subsection{Main Results}

Table~\ref{tab:sea_bed_lang} reports results for small ($\leq$300M) and large ($>$300M) models.
For small models, SEA-LION-Embedding-ModernBERT-300M outperforms harrier-oss-v1-270M on average (0.760 vs. 0.753), achieving the best results on 8 of 10 SEA languages, with especially large gains on lower-resource languages such as Lao and Khmer.
Moreover, compared to embeddinggemma-300m, which performs well on high-resource SEA languages such as Indonesian, Thai, and Vietnamese, our model shows stronger performance on resource-constrained SEA languages, including Myanmar, Khmer, and Lao.

For large models, SEA-LION-Embedding-E5-Large-600M achieves the best overall result with an average score of 0.800, improving over multilingual-e5-large-instruct (0.789) and outperforming the much larger Qwen3-Embedding-8B.
It also outperforms multilingual-e5-large-instruct on 9 of 10 languages.
Compared with the second-best model, harrier-oss-v1-0.6b, our model achieves higher average performance, with stronger gains on lower-resource languages, consistent with the small-model group.
%

\subsection{Ablation Study}

We conduct an ablation study to answer our research questions: \textbf{RQ1:} Data Composition, \textbf{RQ2:} Objective Design, and \textbf{RQ3:} Base Model.
These three factors capture the main levers that practitioners can realistically control when developing embedding models: \textbf{data} determines language coverage and task diversity, the \textbf{objective} determines what geometry the model is trained to preserve, and the \textbf{base encoder} determines the starting multilingual capacity and the achievable performance, efficiency trade-off. 
%

\begin{table}[h!]
\centering
\vspace{-1mm}
\fontsize{7pt}{13pt}
\selectfont
\setlength{\tabcolsep}{2.5pt}
\scalebox{0.9}{
\begin{tabular}{lcc}
\hline
\multicolumn{1}{c|}{\multirow{2}{*}{\textbf{Setup}}}                                                & \multicolumn{2}{c}{\textbf{SEA-BED}} \\ \cline{2-3} 
\multicolumn{1}{l|}{}                                  & \multicolumn{1}{c|}{\textbf{Lang-Avg}} & \textbf{Task-Avg} \\ \hline \hline
\multicolumn{1}{l|}{SEA-LION-Embedding-E5-Large $[$Before: 44.29, 37.6$]$}            & \multicolumn{1}{c|}{\textbf{0.800}}                  &  \textbf{0.754}                 \\ \hline
\multicolumn{3}{l}{\textit{Dataset choices}}                                                                  \\ \hline
\multicolumn{1}{l|}{Use only general datasets}         & \multicolumn{1}{c|}{0.714}                  & 0.688                   \\ 
\multicolumn{1}{l|}{Use only instruction datasets}     & \multicolumn{1}{c|}{0.733}                  & 0.701                  \\ 
\multicolumn{1}{l|}{Use only SEA instruction datasets} & \multicolumn{1}{c|}{0.755}                  & 0.721                  \\ \hline

\multicolumn{3}{l}{$\mathcal{L}_\text{scl}$}                                                                  \\ \hline
\multicolumn{1}{l|}{w/o focal weight} & \multicolumn{1}{c|}{0.792}                  & 0.750                  \\ 
\multicolumn{1}{l|}{Change to vanilla CL~\cite{gao-etal-2021-simcse}}            & \multicolumn{1}{c|}{0.750}                  & 0.699                  \\ \hline
\multicolumn{3}{l}{$\mathcal{L}_\text{sdm}$}                                                                 \\ \hline
\multicolumn{1}{l|}{w/o $\mathcal{L}_\text{sdm}$} & \multicolumn{1}{c|}{0.776}                  &  0.728                 \\ 
\multicolumn{1}{l|}{Change to L2-KD~\cite{reimers-gurevych-2020-making}}                     & \multicolumn{1}{c|}{0.720}                  & 0.672                   \\
\multicolumn{1}{l|}{Change to CKD~\cite{zhao2026kalmembeddingv}}                     & \multicolumn{1}{c|}{0.775}                  &  0.724                 \\ 
\multicolumn{1}{l|}{Change the teacher to Qwen3-Embedding-8B}                     & \multicolumn{1}{c|}{0.778}                  & 0.734                 \\ 
\multicolumn{1}{l|}{$\tau^{t}$ \textless{} $\tau^{s}$} & \multicolumn{1}{c|}{0.781}                  & 0.745                  \\ \hline 
\multicolumn{3}{l}{\textit{Change base models to other models (600M parameters)}}                                                                 \\ \hline
\multicolumn{1}{l|}{Change to XLMR-Large [Before: 0.425, 0.343]}                     & \multicolumn{1}{c|}{0.791}                  & 0.748                  \\
\multicolumn{1}{l|}{Change to mmBERT-base [Before: 0.457, 0.386]}                     & \multicolumn{1}{c|}{0.751}                  & 0.716                  \\ 
\multicolumn{1}{l|}{\begin{tabular}[c]{@{}l@{}}Change to SL-MB-base $[$Before: 0.433, 0.377$]$\end{tabular}}                     & \multicolumn{1}{c|}{0.784}                  & 0.745                   \\

\hline \hline
\end{tabular}}
\vspace{-3mm}
\caption{The ablation study to answer our three RQs. \emph{[Before]} indicates pre-finetuning performance.}
\label{tab:ablation}
\vspace{-3mm}
\end{table}


\subsubsection{Dataset Analysis}
Table~\ref{tab:ablation} shows that using only general or instruction datasets substantially reduces performance; both are needed for robust text embeddings.
Moreover, removing non-SEA-language samples from the instruction datasets causes a large drop from 0.800 to 0.755.
This suggests that non-SEA instruction data provides complementary task and knowledge diversity; for example, MIRACL samples in Thai and Arabic capture different contexts, and training on both improves robustness in downstream tasks.
Overall, robust SEA text embeddings benefit from both SEA-specific data and broader multilingual instruction data.

\subsubsection{Training Objective Analysis}
The training objective is critical to the performance, especially $\mathcal{L}_\text{sdm}$.
Removing $\mathcal{L}_\text{sdm}$ reduces scores from 0.800 to 0.776 in Lang-Avg and from 0.754 to 0.728 in Task-Avg.
Replacing $\mathcal{L}_\text{sdm}$ with alternative KD losses also consistently degrades performance, including L2-KD (0.720) and CKD (0.775).
Replacing either $\mathcal{L}_\text{scl}$ or $\mathcal{L}_\text{sdm}$ with standard alternatives consistently degrades performance.
The temperature design also matters: when using $\tau^t < \tau^s$ instead of $\tau^s < \tau^t$, performance drops to 0.781, suggesting that a sharper student distribution is important for learning a stronger embedding space.

\subsubsection{Model Architecture Analysis}
We further examine whether our pipeline remains effective across base models with different pretraining conditions.
Across all tested bases, including MLM- and contrastive-pretrained initializations, we observe substantial improvements over their starting models, with gains of +0.294 to +0.366 on Lang-Avg and +0.330 to +0.405 on Task-Avg.
E5-Large achieves the best final performance, while XLMR-Large remains competitive, showing that the pipeline is not tied to a single pretraining setting.
These results suggest that strong SEA text embeddings require not only multilingual coverage but also a training pipeline that transfers across diverse base-model initializations.

\subsection{Synthesis: Reproducible Recipe}
\label{sec:synthesis}

Our ablations tell a simple story (Table~\ref{tab:ablation}): robust SEA embeddings require \textbf{both} region-focused coverage and multilingual instruction supervision, and the largest gains come from the \textbf{training objective}, especially $\mathcal{L}_\text{sdm}$ and its temperature. While the base encoder sets the ceiling, the same recipe consistently improves diverse initializations.


%
%

\vspace{-1mm}

\section{Conclusion}
\vspace{-0.5mm}

We present SEA-LION-Embedding, an open, reproducible text-embedding framework for Southeast Asian languages trained only on public data.
Controlled studies of \emph{data composition, training objective, and base-encoder initialization} yield a simple recipe: pair region-focused coverage with multilingual instruction supervision, prioritize the objective (especially $\mathcal{L}_\text{sdm}$ and temperature), and reuse the same procedure across pretrained bases, achieving state-of-the-art SEA-BED results.
Although we focus on SEA, the pipeline is region-agnostic and can be adapted to other underserved regions.

\section*{Limitation}

Similar to previous text embedding works, our study does not exhaustively cover all languages in SEA and training settings relevant to a specific domain task, e.g., adding financial datasets for financial text embedding.
Although we evaluate SEA-LION-Embedding on a large-scale benchmark spanning multiple SEA languages and tasks, there remain many underexplored settings, such as dialectal variation, domain-specific retrieval, and other region-specific downstream applications.
As this paper focuses on building a robust and reproducible SEA text embedding model, we leave broader evaluation challenges for future work on benchmarking and evaluation in Southeast Asian languages.
Moreover, our analysis focuses on the effects of training data, objectives, and backbone models, but does not fully examine other factors such as data scaling laws, annotation quality, or inference-time trade-offs.
These remain important directions for future work.

\section*{Acknowledgments}
This project is supported by the National Research Foundation, Singapore under its National Large Language Models Funding Initiative. Any opinions, findings and conclusions or recommendations expressed in this material are those of the author(s) and do not reflect the views of National Research Foundation, Singapore.

\bibliography{custom}

\clearpage

\appendix

\section*{Appendix}

\section{Hyperparameters} \label{sec:hyperparameters}
The hyperparameters are posted in Table~\ref{tab:hyperparameters}.
Note that we train all models using 64 of H200, where the first step (general datasets using SCL) requires $\sim$6 Hrs, the second step (instruction datasets using SCL) requires $\sim$2 Hrs, and the last step (instruction datasets using SDM) requires $\sim$3 Hrs, for SEA-LION-Embedding-E5-Large-600M.
In addition, the training speed of SEA-LION-Embedding-MondernBERT-300M is half that of SEA-LION-Embedding-E5-Large-600M.
For the inference cost, all models in the same size group have the same inference speed since we did not change any model architectures. 
We will include all dependencies and library versions in our model card for reproducibility. 

\begin{table}[h!]
\centering
\vspace{-3mm}
\fontsize{7pt}{13pt}
\selectfont
\scalebox{.7}{
\begin{tabular}{l|cc}
\hline
\multicolumn{1}{c|}{\multirow{2}{*}{\textbf{Hyperparameter}}} &
  \multicolumn{2}{c}{\textbf{Value}} \\ \cline{2-3} 
\multicolumn{1}{c|}{} &
  \multicolumn{1}{c|}{$\mathcal{L}_\text{scl}$} &
  {$\mathcal{L}_\text{sdm}$} \\ \hline
batch size &
  \multicolumn{1}{c|}{\begin{tabular}[c]{@{}c@{}}600M: 96\\ 300M: 160\end{tabular}} &
  \begin{tabular}[c]{@{}c@{}}600M: 20\\ 300M: 42\end{tabular} \\ \hline
LR             & \multicolumn{2}{c}{1e-5}         \\ \hline
Ep             & \multicolumn{2}{c}{1}            \\ \hline
Temperature(s) &
  \multicolumn{1}{c|}{0.03} &
  \begin{tabular}[c]{@{}c@{}}$\tau^{t}$:0.07\\ $\tau^{s}$:0.03\end{tabular} \\ \hline
LR\_scheduler  & \multicolumn{2}{c}{Cosine}       \\ \hline
$\gamma$       & \multicolumn{1}{c|}{0.5}    & -   \\ \hline
Context length & \multicolumn{2}{c}{512}          \\ \hline
Warm-up steps  & \multicolumn{2}{c}{10\%}         \\ \hline
Pooling        & \multicolumn{2}{c}{Mean pooling} \\ \hline
Optimizer      & \multicolumn{2}{c}{AdamW}        \\ \hline
Precision      & \multicolumn{2}{c}{bf16}         \\ \hline
Teacher model &
  \multicolumn{1}{c|}{-} &
  multilingual-e5-large-instruct \\ \hline
\end{tabular}}
\vspace{-3mm}
\caption{Hyperparameters in this paper.}
\label{tab:hyperparameters}
\vspace{-5mm}
\end{table}

\section{Datasets Details} \label{sec:datasets}

\subsection{Data Leakage}

We want to emphasize that there is no data leakage or using any test set in our model; we did check by scanning through all samples in SEA-BED and our training dataset and found some duplicate examples, such as a question from our training dataset: ``When is the deadline of this question?'' is duplicated with a question in the iAPP-SQuAD dataset. 
However, the document or answer to that question is not the same since it is a different dataset, just a question that is too simple and easy to duplicate.

\subsection{Datasets}

We summarize the full datasets in our work in Tables~\ref{tab:general_dataset} and \ref{tab:instruction_datasets}.
%
%
Comparing with previous work, the pre-training datasets were significantly different in terms of datasets and numbers of samples: \citet{wang2024multilinguale5textembeddings} used more than 1B text pairs, while our work used only 490 M.
On the other hand, \citet{zhao2026kalmembeddingv} proposes a small pre-training dataset with only 46M text pairs.
However, using a small pre-training dataset yields poor results, leading to low performance on SEA languages.
For the downstream or instruction datasets, we found that most recent works used closed datasets~\cite{wang2024multilinguale5textembeddings,wang-etal-2024-improving-text,zhang2025qwen3embeddingadvancingtext,embedding_gemma_2025}, making it hard to compare the size of the training dataset. 

\begin{table}[h!]
\centering
\vspace{-3mm}
\fontsize{7pt}{13pt}
\selectfont
\scalebox{0.9}{
\begin{tabular}{l|cccc}
\hline
\multicolumn{1}{c|}{\textbf{Dataset}} &
  \multicolumn{1}{c|}{\textbf{Type}} &
  \multicolumn{1}{c|}{\textbf{Categ.}} &
  \multicolumn{1}{c|}{\textbf{Language}} &
  \textbf{Pairs} \\ \hline
\href{https://huggingface.co/datasets/HIT-TMG/KaLM-embedding-pretrain-data}{kalm\_pretraining} &
  \multicolumn{1}{c|}{Retrieval} &
  \multicolumn{1}{c|}{s2p} &
  \multicolumn{1}{c|}{English} &
  {\color[HTML]{434343} 23,670,898} \\ 
\href{https://huggingface.co/datasets/HuggingFaceFW/finetranslations}{finetranslation}     & \multicolumn{1}{c|}{BitextMining} & \multicolumn{1}{c|}{s2s} & \multicolumn{1}{c|}{SEA}     & {\color[HTML]{434343} 61,263,694} \\ 
\href{https://huggingface.co/datasets/wikimedia/wikipedia}{wikipedia}           & \multicolumn{1}{c|}{Retrieval}    & \multicolumn{1}{c|}{s2s} & \multicolumn{1}{c|}{SEA}     & {\color[HTML]{434343} 4,039,056}  \\ 
\href{https://huggingface.co/datasets/allenai/c4}{c4}                  & \multicolumn{1}{c|}{Retrieval}    & \multicolumn{1}{c|}{s2s} & \multicolumn{1}{c|}{SEA}     & {\color[HTML]{434343} 51,711,918} \\ 
\href{https://huggingface.co/datasets/sentence-transformers/s2orc}{s2orc}               & \multicolumn{1}{c|}{Retrieval}    & \multicolumn{1}{c|}{s2p} & \multicolumn{1}{c|}{English} & {\color[HTML]{434343} 41,769,185} \\ 
\href{https://huggingface.co/datasets/sentence-transformers/parallel-sentences-tatoeba}{tatoeba}             & \multicolumn{1}{c|}{BitextMining} & \multicolumn{1}{c|}{s2s} & \multicolumn{1}{c|}{SEA}     & {\color[HTML]{434343} 70,215}     \\ 
\href{https://huggingface.co/datasets/sentence-transformers/parallel-sentences-talks}{talks}               & \multicolumn{1}{c|}{BitextMining} & \multicolumn{1}{c|}{s2s} & \multicolumn{1}{c|}{SEA}     & {\color[HTML]{434343} 704,527}    \\ 
\href{https://huggingface.co/datasets/sentence-transformers/parallel-sentences-jw300}{jw300}               & \multicolumn{1}{c|}{BitextMining} & \multicolumn{1}{c|}{s2s} & \multicolumn{1}{c|}{SEA}     & {\color[HTML]{434343} 2,772,088}  \\ 
\href{https://huggingface.co/datasets/sentence-transformers/parallel-sentences-wikimatrix}{wikimatrix}          & \multicolumn{1}{c|}{BitextMining} & \multicolumn{1}{c|}{s2s} & \multicolumn{1}{c|}{SEA}     & {\color[HTML]{434343} 842,334}    \\ 
\href{https://huggingface.co/datasets/sentence-transformers/parallel-sentences-opensubtitles}{opensubtitles}       & \multicolumn{1}{c|}{BitextMining} & \multicolumn{1}{c|}{s2s} & \multicolumn{1}{c|}{SEA}     & {\color[HTML]{434343} 10,473,771} \\ 
\href{https://huggingface.co/datasets/sentence-transformers/parallel-sentences-opus-100}{opus-100}            & \multicolumn{1}{c|}{BitextMining} & \multicolumn{1}{c|}{s2s} & \multicolumn{1}{c|}{SEA}     & {\color[HTML]{434343} 5,136,077}  \\ 
\href{https://huggingface.co/datasets/sentence-transformers/parallel-sentences-ccmatrix}{ccmatrix}            & \multicolumn{1}{c|}{Retrieval}    & \multicolumn{1}{c|}{s2s} & \multicolumn{1}{c|}{SEA}     & {\color[HTML]{434343} 43,113,828} \\ \hline
\textbf{Total size} & \multicolumn{4}{c}{\textbf{245,567,591}}                                                                                       \\ \hline
\end{tabular}}
\vspace{-3mm}
\caption{The general text datasets, where no instructions were included in these datasets.}
\label{tab:general_dataset}
\vspace{-5mm}
\end{table}

\subsection{Instructions}

To induce task-specific instruction-following behavior in the embeddings, we prepend a task instruction to each query. Formally, the instructed query is written as:

\vspace{-3mm}
\begin{equation}
\small
\text{Input} = \texttt{Instruct: \{task instruction\} Query: } q .
\end{equation}

We add instructions to all samples in Table~\ref{tab:instruction_datasets}, where we applied $\texttt{\{task instruction\}}$ for different task types from Table~\ref{tab:instructions}.
For both asymmetric and symmetric tasks, in contrast to previous works~\cite{zhao2026kalmembeddingv,embedding_gemma_2025}, we use the same instruction without adding additional prepends to the instruction.
We found that it is simpler, easier to reproduce, and does not affect any downstream task performances.
%

\begin{table}[h!]
\centering
\fontsize{7pt}{13pt}
\selectfont
\scalebox{0.8}{
\begin{tabular}{l|l}
\hline
\multicolumn{1}{c|}{\textbf{Task}} &
  \multicolumn{1}{c}{\textbf{Instruction}} \\ \hline
BitextMining &
  Retrieve parallel sentences \\ \hline
Retrieval &
  \begin{tabular}[c]{@{}l@{}}Given a passage that is guaranteed to contain the answer, \\ retrieve relevant passages that answer the query\end{tabular} \\ \hline
PairClassification &
  Retrieve semantically similar text \\ \hline
Classification &
  Classify text into its appropriate category \\ \hline
STS &
  Retrieve semantically similar text \\ \hline
InstructionRetrieval &
  \begin{tabular}[c]{@{}l@{}}Given a instruction and a output, \\ retrieve the most relevant output that answer the instruction\end{tabular} \\ \hline
\end{tabular}}
\vspace{-3mm}
\caption{The instruction examples.}
\label{tab:instructions}
\vspace{-5mm}
\end{table}

\begin{table*}[h!]
\centering
\fontsize{7pt}{13pt}
\selectfont
\scalebox{.7}{
\begin{tabular}{lcccccccccc}
\hline
\multicolumn{1}{l|}{\textbf{Model}} &
  \multicolumn{1}{c|}{\textbf{STS}} &
  \multicolumn{1}{c|}{\textbf{Classification}} &
  \multicolumn{1}{c|}{\textbf{PairClassification}} &
  \multicolumn{1}{c|}{\textbf{QARetrieval}} &
  \multicolumn{1}{c|}{\textbf{InstructionRetrieval}} &
  \multicolumn{1}{c|}{\textbf{BitextMining}} &
  \multicolumn{1}{c|}{\textbf{MultiLabelClassification}} &
  \multicolumn{1}{c|}{\textbf{Reranking}} &
  \multicolumn{1}{c|}{\textbf{Clustering}} &
  \textbf{Avg.} \\ \hline \hline
  \multicolumn{11}{l}{\textit{$\leq$300M parameters}} \\ \hline

\multicolumn{1}{l|}{embeddinggemma-300m} &
  \multicolumn{1}{c|}{0.681} &
  \multicolumn{1}{c|}{\textbf{0.772}} &
  \multicolumn{1}{c|}{\textbf{0.696}} &
  \multicolumn{1}{c|}{\textbf{0.771}} &
  \multicolumn{1}{c|}{\textbf{0.707}} &
  \multicolumn{1}{c|}{0.729} &
  \multicolumn{1}{c|}{0.892} &
  \multicolumn{1}{c|}{\textbf{0.784}} &
  \multicolumn{1}{c|}{0.402} &
  0.715 \\ 
\multicolumn{1}{l|}{harrier-oss-v1-270m} &
  \multicolumn{1}{c|}{0.712} &
  \multicolumn{1}{c|}{0.753} &
  \multicolumn{1}{c|}{0.667} &
  \multicolumn{1}{c|}{0.745} &
  \multicolumn{1}{c|}{0.625} &
  \multicolumn{1}{c|}{0.831} &
  \multicolumn{1}{c|}{0.883} &
  \multicolumn{1}{c|}{0.695} &
  \multicolumn{1}{c|}{0.446} &
  0.706 \\ 
  \multicolumn{1}{l|}{SEA-LION-Embedding-ModernBERT-300M} &
  \multicolumn{1}{c|}{\textbf{0.739}} &
  \multicolumn{1}{c|}{0.758} &
  \multicolumn{1}{c|}{0.661} &
  \multicolumn{1}{c|}{0.736} &
  \multicolumn{1}{c|}{0.639} &
  \multicolumn{1}{c|}{\textbf{0.862}} &
  \multicolumn{1}{c|}{\textbf{0.894}} &
  \multicolumn{1}{c|}{0.726} &
  \multicolumn{1}{c|}{\textbf{0.493}} &
  \textbf{0.723$^\dagger$} \\  \hline
\multicolumn{11}{l}{\textit{$>$300M parameters}} \\ \hline


\multicolumn{1}{l|}{multilingual-e5-large-instruct} &
  \multicolumn{1}{c|}{0.756} &
  \multicolumn{1}{c|}{0.777} &
  \multicolumn{1}{c|}{0.666} &
  \multicolumn{1}{c|}{0.772} &
  \multicolumn{1}{c|}{0.691} &
  \multicolumn{1}{c|}{0.879} &
  \multicolumn{1}{c|}{0.878} &
  \multicolumn{1}{c|}{0.772} &
  \multicolumn{1}{c|}{{0.581}} &
  0.752 \\ 
\multicolumn{1}{l|}{Qwen3-Embedding-0.6B} &
  \multicolumn{1}{c|}{0.657} &
  \multicolumn{1}{c|}{0.745} &
  \multicolumn{1}{c|}{0.604} &
  \multicolumn{1}{c|}{{0.762}} &
  \multicolumn{1}{c|}{{0.658}} &
  \multicolumn{1}{c|}{0.565} &
  \multicolumn{1}{c|}{{0.882}} &
  \multicolumn{1}{c|}{{0.750}} &
  \multicolumn{1}{c|}{0.439} &
  0.674 \\ 
\multicolumn{1}{l|}{Qwen3-Embedding-8B} &
  \multicolumn{1}{c|}{0.753} &
  \multicolumn{1}{c|}{\textbf{0.786}} &
  \multicolumn{1}{c|}{0.631} &
  \multicolumn{1}{c|}{\textbf{0.820}} &
  \multicolumn{1}{c|}{{0.708}} &
  \multicolumn{1}{c|}{0.848} &
  \multicolumn{1}{c|}{\textbf{0.906}} &
  \multicolumn{1}{c|}{\textbf{0.785}} &
  \multicolumn{1}{c|}{0.529} &
  0.752 \\ 
\multicolumn{1}{l|}{bge-m3 (Dense)} &
  \multicolumn{1}{c|}{0.733} &
  \multicolumn{1}{c|}{0.760} &
  \multicolumn{1}{c|}{0.687} &
  \multicolumn{1}{c|}{0.736} &
  \multicolumn{1}{c|}{0.585} &
  \multicolumn{1}{c|}{0.862} &
  \multicolumn{1}{c|}{0.899} &
  \multicolumn{1}{c|}{0.760} &
  \multicolumn{1}{c|}{0.422} &
  0.716 \\ 
\multicolumn{1}{l|}{Cohere-embed-multilingual-v3.0} &
  \multicolumn{1}{c|}{0.731} &
  \multicolumn{1}{c|}{0.785} &
  \multicolumn{1}{c|}{0.661} &
  \multicolumn{1}{c|}{0.782} &
  \multicolumn{1}{c|}{0.656} &
  \multicolumn{1}{c|}{0.883} &
  \multicolumn{1}{c|}{0.900} &
  \multicolumn{1}{c|}{0.778} &
  \multicolumn{1}{c|}{0.490} &
  0.741 \\ 
\multicolumn{1}{l|}{jina-embeddings-v5} &
  \multicolumn{1}{c|}{0.730} &
  \multicolumn{1}{c|}{0.729} &
  \multicolumn{1}{c|}{0.533} &
  \multicolumn{1}{c|}{0.683} &
  \multicolumn{1}{c|}{\textbf{0.855}} &
  \multicolumn{1}{c|}{0.632} &
  \multicolumn{1}{c|}{0.765} &
  \multicolumn{1}{c|}{0.767} &
  \multicolumn{1}{c|}{\textbf{0.705}} &
  0.711 \\ 
\multicolumn{1}{l|}{harrier-oss-v1-0.6b} &
  \multicolumn{1}{c|}{0.740} &
  \multicolumn{1}{c|}{0.772} &
  \multicolumn{1}{c|}{\textbf{0.691}} &
  \multicolumn{1}{c|}{0.773} &
  \multicolumn{1}{c|}{0.685} &
  \multicolumn{1}{c|}{0.870} &
  \multicolumn{1}{c|}{0.890} &
  \multicolumn{1}{c|}{0.737} &
  \multicolumn{1}{c|}{0.454} &
  0.735 \\ 
\multicolumn{1}{l|}{SEA-LION-Embedding-E5-Large-600M} &
  \multicolumn{1}{c|}{\textbf{0.765}} &
  \multicolumn{1}{c|}{0.761} &
  \multicolumn{1}{c|}{0.673} &
  \multicolumn{1}{c|}{0.778} &
  \multicolumn{1}{c|}{0.700} &
  \multicolumn{1}{c|}{\textbf{0.896}} &
  \multicolumn{1}{c|}{0.894} &
  \multicolumn{1}{c|}{0.781} &
  \multicolumn{1}{c|}{0.540} &
  \textbf{0.754} \\ \hline \hline
\end{tabular}}
\vspace{-3mm}
\caption{The performance of all models on SEA-BED using the task average metric. In addition, $\dagger$ represents the significant improvement from the previous SOTA in each group using McNemar's test.}
\label{tab:sea_bed_task}
\vspace{-2mm}
\end{table*}

\begin{table*}[h!]
\centering
\fontsize{7pt}{13pt}
\selectfont
\scalebox{.7}{
\begin{tabular}{llllllllll}
\hline
\multicolumn{1}{l|}{\textbf{Model}}                     & \multicolumn{1}{l|}{\textbf{Classification}} & \multicolumn{1}{l|}{\textbf{Clustering}} & \multicolumn{1}{l|}{\textbf{Pair Classification}} & \multicolumn{1}{l|}{\textbf{Reranking}} & \multicolumn{1}{l|}{\textbf{Retrieval}} & \multicolumn{1}{l|}{\textbf{STS}}   & \multicolumn{1}{l|}{\textbf{Summarization}} & \multicolumn{1}{l|}{\textbf{Mean (Task)}} & \textbf{Mean (Task Type)} \\ \hline
\multicolumn{10}{l}{\textit{$\leq$300M parameters}}                                                                                                                                                                                                                                                                                                                                                                                             \\ \hline
\multicolumn{1}{l|}{embeddinggemma-300m}                & \multicolumn{1}{l|}{\textbf{0.875}}          & \multicolumn{1}{l|}{\textbf{0.533}}      & \multicolumn{1}{l|}{\textbf{0.848}}               & \multicolumn{1}{l|}{\textbf{0.469}}     & \multicolumn{1}{l|}{\textbf{0.546}}     & \multicolumn{1}{l|}{\textbf{0.829}} & \multicolumn{1}{l|}{\textbf{0.378}}         & \multicolumn{1}{l|}{\textbf{0.684}}       & \textbf{0.640}            \\
\multicolumn{1}{l|}{harrier-oss-v1-270m}                & \multicolumn{1}{l|}{0.765}                   & \multicolumn{1}{l|}{0.472}               & \multicolumn{1}{l|}{0.812}                        & \multicolumn{1}{l|}{0.444}              & \multicolumn{1}{l|}{0.410}              & \multicolumn{1}{l|}{0.766}          & \multicolumn{1}{l|}{0.240}                  & \multicolumn{1}{l|}{0.596}                & 0.558                     \\
\multicolumn{1}{l|}{SEA-LION-ModernBERT-Embedding-300M} & \multicolumn{1}{l|}{0.776}                   & \multicolumn{1}{l|}{0.484}               & \multicolumn{1}{l|}{0.832}                        & \multicolumn{1}{l|}{0.446}              & \multicolumn{1}{l|}{0.448}              & \multicolumn{1}{l|}{0.818}          & \multicolumn{1}{l|}{0.270}                  & \multicolumn{1}{l|}{0.624}                & 0.582                     \\ \hline
\multicolumn{10}{l}{\textit{$>$300M parameters}}                                                                                                                                                                                                                                                                                                                                                                                           \\ \hline
\multicolumn{1}{l|}{multilingual-e5-large-instruct}     & \multicolumn{1}{l|}{0.753}                   & \multicolumn{1}{l|}{0.517}               & \multicolumn{1}{l|}{0.862}                        & \multicolumn{1}{l|}{0.464}              & \multicolumn{1}{l|}{0.536}              & \multicolumn{1}{l|}{0.846}          & \multicolumn{1}{l|}{0.306}                  & \multicolumn{1}{l|}{0.657}                & 0.612                     \\
\multicolumn{1}{l|}{Qwen3-Embedding-8B}                 & \multicolumn{1}{l|}{\textbf{0.904}}          & \multicolumn{1}{l|}{\textbf{0.583}}      & \multicolumn{1}{l|}{\textbf{0.880}}               & \multicolumn{1}{l|}{\textbf{0.509}}     & \multicolumn{1}{l|}{\textbf{0.676}}     & \multicolumn{1}{l|}{\textbf{0.887}} & \multicolumn{1}{l|}{\textbf{0.352}}         & \multicolumn{1}{l|}{\textbf{0.748}}       & \textbf{0.685}            \\
\multicolumn{1}{l|}{bge-m3 (Dense)}                     & \multicolumn{1}{l|}{0.774}                   & \multicolumn{1}{l|}{0.407}               & \multicolumn{1}{l|}{0.845}                        & \multicolumn{1}{l|}{0.460}              & \multicolumn{1}{l|}{0.496}              & \multicolumn{1}{l|}{0.809}          & \multicolumn{1}{l|}{0.348}                  & \multicolumn{1}{l|}{0.622}                & 0.591                     \\
\multicolumn{1}{l|}{jina-embeddings-v3}                 & \multicolumn{1}{l|}{0.854}                   & \multicolumn{1}{l|}{0.480}               & \multicolumn{1}{l|}{0.840}                        & \multicolumn{1}{l|}{0.480}              & \multicolumn{1}{l|}{0.526}              & \multicolumn{1}{l|}{0.857}          & \multicolumn{1}{l|}{0.329}                  & \multicolumn{1}{l|}{0.669}                & 0.623                     \\
\multicolumn{1}{l|}{harrier-oss-v1-0.6b}                & \multicolumn{1}{l|}{0.883}                   & \multicolumn{1}{l|}{0.566}               & \multicolumn{1}{l|}{0.867}                        & \multicolumn{1}{l|}{0.470}              & \multicolumn{1}{l|}{0.586}              & \multicolumn{1}{l|}{0.851}          & \multicolumn{1}{l|}{0.315}                  & \multicolumn{1}{l|}{0.706}                & 0.648                     \\
\multicolumn{1}{l|}{SEA-LION-E5-Embedding-600M}         & \multicolumn{1}{l|}{0.778}                   & \multicolumn{1}{l|}{0.511}               & \multicolumn{1}{l|}{0.874}                        & \multicolumn{1}{l|}{0.471}              & \multicolumn{1}{l|}{0.502}              & \multicolumn{1}{l|}{0.848}          & \multicolumn{1}{l|}{0.316}                  & \multicolumn{1}{l|}{0.655}                & 0.614 \\ \hline \hline                    
\end{tabular}}
\vspace{-3mm}
\caption{The performance of all models on MTEB.}
\label{tab:mteb_task}
\vspace{-2mm}
\end{table*}

\begin{table*}[h!]
\centering
\fontsize{7pt}{13pt}
\selectfont
\scalebox{.7}{
\begin{tabular}{lllllllll}
\hline
\multicolumn{1}{l|}{\textbf{Model}}                     & \multicolumn{1}{l|}{\textbf{Classification}} & \multicolumn{1}{l|}{\textbf{Clustering}} & \multicolumn{1}{l|}{\textbf{Pair Classification}} & \multicolumn{1}{l|}{\textbf{Reranking}} & \multicolumn{1}{l|}{\textbf{Retrieval}} & \multicolumn{1}{l|}{\textbf{STS}}   & \multicolumn{1}{l|}{\textbf{Mean (Task)}} & \textbf{Mean (Task Type)} \\ \hline
\multicolumn{9}{l}{\textit{$\leq$300M parameters}}                                                                                                                                                                                                                                                                                                                                                \\ \hline
\multicolumn{1}{l|}{embeddinggemma-300m}                & \multicolumn{1}{l|}{\textbf{0.725}}          & \multicolumn{1}{l|}{0.504}               & \multicolumn{1}{l|}{\textbf{0.733}}               & \multicolumn{1}{l|}{0.589}              & \multicolumn{1}{l|}{\textbf{0.642}}     & \multicolumn{1}{l|}{\textbf{0.469}} & \multicolumn{1}{l|}{\textbf{0.600}}       & \textbf{0.610}            \\
\multicolumn{1}{l|}{harrier-oss-v1-270m}                & \multicolumn{1}{l|}{0.696}                   & \multicolumn{1}{l|}{0.494}               & \multicolumn{1}{l|}{0.606}                        & \multicolumn{1}{l|}{\textbf{0.621}}     & \multicolumn{1}{l|}{0.589}              & \multicolumn{1}{l|}{0.403}          & \multicolumn{1}{l|}{0.561}                & 0.568                     \\
\multicolumn{1}{l|}{SEA-LION-ModernBERT-Embedding-300M} & \multicolumn{1}{l|}{0.713}                   & \multicolumn{1}{l|}{\textbf{0.512}}      & \multicolumn{1}{l|}{0.650}                        & \multicolumn{1}{l|}{0.563}              & \multicolumn{1}{l|}{0.590}              & \multicolumn{1}{l|}{0.464}          & \multicolumn{1}{l|}{0.576}                & 0.582                     \\ \hline
\multicolumn{9}{l}{\textit{$>$300M parameters}}                                                                                                                                                                                                                                                                                                                                              \\ \hline
\multicolumn{1}{l|}{multilingual-e5-large-instruct}     & \multicolumn{1}{l|}{0.706}                   & \multicolumn{1}{l|}{0.532}               & \multicolumn{1}{l|}{0.668}                        & \multicolumn{1}{l|}{0.611}              & \multicolumn{1}{l|}{0.642}              & \multicolumn{1}{l|}{0.482}          & \multicolumn{1}{l|}{0.602}                & 0.607                     \\
\multicolumn{1}{l|}{Qwen3-Embedding-8B}                 & \multicolumn{1}{l|}{\textbf{0.771}}          & \multicolumn{1}{l|}{\textbf{0.799}}      & \multicolumn{1}{l|}{\textbf{0.840}}               & \multicolumn{1}{l|}{\textbf{0.682}}     & \multicolumn{1}{l|}{\textbf{0.781}}     & \multicolumn{1}{l|}{\textbf{0.634}} & \multicolumn{1}{l|}{\textbf{0.739}}       & \textbf{0.751}            \\
\multicolumn{1}{l|}{bge-m3 (Dense)}                     & \multicolumn{1}{l|}{0.743}                   & \multicolumn{1}{l|}{0.536}               & \multicolumn{1}{l|}{0.682}                        & \multicolumn{1}{l|}{0.620}              & \multicolumn{1}{l|}{0.557}              & \multicolumn{1}{l|}{0.508}          & \multicolumn{1}{l|}{0.595}                & 0.608                     \\
\multicolumn{1}{l|}{jina-embeddings-v3}                 & \multicolumn{1}{l|}{0.723}                   & \multicolumn{1}{l|}{0.502}               & \multicolumn{1}{l|}{0.669}                        & \multicolumn{1}{l|}{0.606}              & \multicolumn{1}{l|}{0.686}              & \multicolumn{1}{l|}{0.525}          & \multicolumn{1}{l|}{0.621}                & 0.618                     \\
\multicolumn{1}{l|}{harrier-oss-v1-0.6b}                & \multicolumn{1}{l|}{0.754}                   & \multicolumn{1}{l|}{0.647}               & \multicolumn{1}{l|}{0.712}                        & \multicolumn{1}{l|}{0.656}              & \multicolumn{1}{l|}{0.713}              & \multicolumn{1}{l|}{0.501}          & \multicolumn{1}{l|}{0.657}                & 0.664                     \\
\multicolumn{1}{l|}{SEA-LION-E5-Embedding-600M}         & \multicolumn{1}{l|}{0.710}                   & \multicolumn{1}{l|}{0.527}               & \multicolumn{1}{l|}{0.666}                        & \multicolumn{1}{l|}{0.607}              & \multicolumn{1}{l|}{0.650}              & \multicolumn{1}{l|}{0.488}          & \multicolumn{1}{l|}{0.605}                & 0.608       \\ \hline \hline             
\end{tabular}}
\vspace{-3mm}
\caption{The performance of all models on CMTEB.}
\label{tab:cmteb_task}
\vspace{-2mm}
\end{table*}

\section{SEA-BED Results} \label{sec:sea_bed_appendix}

We report the task score in Table~\ref{tab:sea_bed_task} using the same models as Table~\ref{tab:sea_bed_lang}.
We found that the performance of our model in the small group significantly outperforms SOTA.
In addition, unlike the language metric in Table~\ref{tab:sea_bed_lang}, some models perform poorly in this metric, such as harrier-oss-v1.
In contrast, our model consistency outperform all models similar to the language metric for both small and large groups.

\section{MTEB and CMTEB Results} \label{sec:mteb_and_cmteb_results}

Since people in the SEA region also widely speak Chinese and English, we report the English (MTEB) and Chinese (CMTEB) results in Tables~\ref{tab:mteb_task} and \ref{tab:cmteb_task} to align with real-world scenarios.
Note that we reproduce all scores using the default settings of MTEB and CMTEB. 
However, we found that, when the codes or settings from the original model are not available for reproducible, we cannot reproduce the same score as reported, making the Reprod. Env. in Table~\ref{tab:overview} marked ``No''. 
This highlights the important of open-science, which makes all papers reproducible.

The experimental results demonstrate that compared to the small model group, our model performs the second-best, only behind embeddinggemma-300m.
For the large model group, our model still performs better than the teacher model, multilingual-e5-large-instruct, in the average score.
In addition, we notice that models that performs poor on SEA-BED achieve a better score on English and Chinese datasets. 
This emphasizes that the model is supported well for English and Chinese, both of which are high-resource languages, but lacks robustness in low-resource languages, especially for SEA languages.

\begin{table*}[h!]
\centering
\vspace{-3mm}
\fontsize{7pt}{13pt}
\selectfont
\scalebox{0.7}{
\begin{tabular}{llllrllllr}
\hline
Source & Type & Categ. & Language & Pairs & Source & Type & Categ. & Language & Pairs \\
\hline
\href{https://huggingface.co/datasets/m-a-p/CodeFeedback-Filtered-Instruction}{CodeFeedback} & Retrieval & s2p & en & 49,090 & \href{https://huggingface.co/datasets/rusano/ELI5_custom}{ELI5} & Retrieval & s2p & en & 76,408 \\
\href{https://github.com/chaitanyamalaviya/ExpertQA}{ExpertQA} & Retrieval & s2p & en & 1,252 & \href{https://github.com/allenai/gooaq}{GooAQ} & Retrieval & s2p & en & 49,833 \\
\href{https://hf.co/datasets/GritLM/MEDI2BGE}{MEDI2BGE} & Retrieval & s2p & en & 71,790 & \href{https://huggingface.co/datasets/Open-Orca/OpenOrca}{OpenOrca} & Retrieval & s2p & en & 38,623 \\
\href{https://huggingface.co/datasets/sentence-transformers/paq}{PAQ} & Retrieval & s2p & en & 49,849 & \href{https://huggingface.co/datasets/qiaojin/PubMedQA}{PubMedQA} & Retrieval & s2p & en & 79,954 \\
\href{https://huggingface.co/datasets/kyunghyuncho/search_qa}{SearchQA} & Retrieval & s2p & en & 9,988 & \href{https://huggingface.co/datasets/TitanMLData/arxiv_qa}{arxiv\_qa} & Retrieval & s2p & en & 17,927 \\
\href{https://huggingface.co/datasets/intfloat/multilingual_cc_news}{CC-News} & Retrieval & s2p & en & 28,246 & \href{https://huggingface.co/datasets/irds/cord19_trec-covid}{TREC-COVID} & Retrieval & s2p & en & 48,517 \\
\href{https://huggingface.co/datasets/BeIR/dbpedia-entity-generated-queries}{DBpedia-Entity} & Retrieval & s2p & en & 96,792 & \href{https://huggingface.co/datasets/tasksource/esci}{ESCI} & Retrieval & s2p & en & 26,043 \\
\href{https://huggingface.co/datasets/maxzoech/fever}{FEVER} & Retrieval & s2p & en & 87,216 & \href{https://huggingface.co/datasets/irds/beir_fiqa_train}{FiQA} & Retrieval & s2p & en & 4,689 \\
\href{https://huggingface.co/datasets/hotpotqa/hotpot_qa}{HotpotQA} & Retrieval & s2p & en & 150,153 & \href{https://huggingface.co/datasets/Shitao/MLDR}{MLDR} & Retrieval & s2p & en & 31,097 \\
\href{https://huggingface.co/datasets/Tevatron/msmarco-passage}{MSMARCO} & Retrieval & s2p & en & 174,190 & \href{https://huggingface.co/datasets/mteb/msmarco-v2}{MSMARCO-v2} & Retrieval & s2p & en & 258,617 \\
\href{https://huggingface.co/datasets/BeIR/nfcorpus-generated-queries}{NFCorpus} & Retrieval & s2p & en & 10,471 & \href{https://huggingface.co/datasets/neural-bridge/rag-dataset-12000}{rag-dataset-12000} & Retrieval & s2p & en & 9,272 \\
\href{https://huggingface.co/datasets/Tevatron/scifact}{SciFact} & Retrieval & s2p & en & 794 & \href{https://huggingface.co/datasets/rajpurkar/squad_v2}{SQuAD 2.0} & Retrieval & s2p & en & 125,816 \\
\href{https://huggingface.co/datasets/multi-train/emb-triviaqa-train}{TriviaQA} & Retrieval & s2p & en & 44,442 & \href{https://huggingface.co/datasets/openai/webgpt_comparisons}{WebGPT Comparisons} & Retrieval & s2p & en & 18,924 \\
\href{https://huggingface.co/datasets/Tevatron/wikipedia-nq}{Natural Questions} & Retrieval & s2p & en & 56,377 & \href{https://huggingface.co/datasets/sentence-transformers/yahoo-answers}{Yahoo Answers} & Retrieval & s2p & en & 21,724 \\
\href{http://nlp.cis.unimelb.edu.au/resources/cqadupstack/}{CQADupStack} & Retrieval & s2p & en & 7,356 & \href{https://huggingface.co/datasets/kiddothe2b/contract-nli}{ContractNLI} & PairClassification & s2s & en & 628 \\
\href{https://huggingface.co/datasets/SetFit/mnli}{MultiNLI} & PairClassification & s2s & en & 63,701 & \href{https://huggingface.co/datasets/breakend/nllb-multi-domain}{NLLB} & STS & s2s & en & 26,504 \\
\href{https://huggingface.co/datasets/sentence-transformers/embedding-training-data}{Quora} & STS & s2s & en & 89,558 & \href{https://huggingface.co/datasets/multi-train/WikiAnswers_1107}{WikiAnswers} & STS & s2s & en & 47,686 \\
\href{https://huggingface.co/datasets/JeremiahZ/simcse_sup_nli}{SimCSE NLI} & PairClassification & s2s & en & 217,099 & \href{https://huggingface.co/datasets/stanfordnlp/snli}{SNLI} & PairClassification & s2s & en & 16,480 \\
\href{https://huggingface.co/datasets/mteb/raw_arxiv}{arXiv} & Classification & s2s, p2s & en & 14,529 & \href{https://huggingface.co/datasets/mteb/raw_biorxiv}{Biorxiv} & Classification & s2s, p2s & en & 6,787 \\
\href{https://huggingface.co/datasets/mteb/raw_medrxiv}{Medrxiv} & Classification & s2s, p2s & en & 1,999 & \href{https://github.com/UKPLab/TWEAC-qa-agent-selection/tree/master/data/reddit/train}{Reddit-Clustering} & Classification & s2s & en & 25,600 \\
\href{https://huggingface.co/datasets/sentence-transformers/reddit-title-body}{Reddit-Clustering-P2P} & Classification & p2s & en & 42,480 & \href{https://github.com/UKPLab/TWEAC-qa-agent-selection/tree/master/data/stackexchange/train}{Stackexchange-Clustering} & Classification & s2s & en & 50,530 \\
\href{https://huggingface.co/datasets/flax-sentence-embeddings/stackexchange_title_body_jsonl}{Stackexchange-Clustering-P2P} & Classification & p2s & en & 48,800 & \href{https://scikit-learn.org/0.19/datasets/twenty_newsgroups.html}{TwentyNewsgroups-Clustering} & Classification & s2s & en & 6,233 \\
\href{https://huggingface.co/datasets/mteb/amazon_polarity}{AmazonPolarity} & Classification & s2s & en & 9,007 & \href{https://huggingface.co/datasets/mteb/imdb}{IMDB} & Classification & s2s & en & 8,575 \\
\href{https://huggingface.co/datasets/mteb/banking77}{banking77} & Classification & s2s & en & 9,937 & \href{https://huggingface.co/datasets/mteb/emotion}{EmotionClassification} & Classification & s2s & en & 10,000 \\
\href{https://huggingface.co/datasets/mteb/tweet_sentiment_extraction}{TweetSentimentExtraction} & Classification & s2s & en & 10,000 & \href{https://huggingface.co/datasets/mteb/toxic_conversations_50k}{ToxicConversations} & Classification & s2s & en & 7,800 \\
\href{https://huggingface.co/datasets/shibing624/AdvertiseGen}{AdvertiseGen} & Retrieval & s2p & zh & 17,526 & \href{https://www.luge.ai/}{CHEF} & Retrieval & s2p & zh & 4,824 \\
\href{https://huggingface.co/datasets/michaelwzhu/ChatMed_Consult_Dataset}{ChatMed-Dataset} & Retrieval & s2p & zh & 18,608 & \href{https://huggingface.co/datasets/erhwenkuo/squad-cmrc2018-zhtw}{CMRC 2018} & Retrieval & s2p & zh & 9,753 \\
\href{https://huggingface.co/datasets/voidful/DRCD}{DRCD} & Retrieval & s2p & zh & 4,714 & \href{https://huggingface.co/datasets/hugcyp/LCSTS}{LCSTS} & Retrieval & s2p & zh & 19,535 \\
\href{https://huggingface.co/datasets/paralym/lima-chinese}{LIMA} & Retrieval & s2p & zh & 1,991 & \href{https://github.com/Alibaba-NLP/Multi-CPR}{Multi-CPR} & Retrieval & s2p & zh & 234,587 \\
\href{https://huggingface.co/datasets/C-MTEB/PAWSX}{PAWS-X (zh)} & Retrieval & s2p & zh & 19,289 & \href{https://github.com/sufengniu/RefGPT/blob/main/README_EN.md}{RefGPT} & Retrieval & s2p & zh & 49,896 \\
\href{https://huggingface.co/datasets/THUIR/T2Ranking}{T2Ranking} & Retrieval & s2p & zh & 188,606 & \href{https://huggingface.co/datasets/SirlyDreamer/THUCNews}{THUCNews} & Retrieval & s2p & zh & 19,288 \\
\href{https://www.luge.ai/}{UMETRIP-QA} & Retrieval & s2p & zh & 2,537 & \href{https://github.com/thunlp/WebCPM}{WebCPM} & Retrieval & s2p & zh & 1,602 \\
\href{https://www.datafountain.cn/competitions/424/datasets}{cCOVID-News} & Retrieval & s2p & zh & 4,727 & \href{https://huggingface.co/datasets/wangrongsheng/cMedQA-V2.0}{cMedQA-V2.0} & Retrieval & s2p & zh & 88,109 \\
\href{https://huggingface.co/datasets/neuclir/csl}{CSL} & Retrieval & s2p & zh & 19,945 & \href{https://huggingface.co/datasets/sentence-transformers/dureader}{DuReader} & Retrieval & s2p & zh & 79,229 \\
\href{https://huggingface.co/datasets/luozhouyang/dureader}{DuReader\_checklist} & Retrieval & s2p & zh & 97,764 & \href{https://huggingface.co/datasets/sentence-transformers/law-gpt}{law-gpt} & Retrieval & s2p & zh & 500 \\
\href{https://www.heywhale.com/mw/dataset/5e953ca8e7ec38002d02fca7/content}{lawzhidao} & Retrieval & s2p & zh & 6,784 & \href{https://huggingface.co/datasets/unicamp-dl/mmarco}{mMARCO (zh)} & Retrieval & s2p & zh & 379,870 \\
\href{https://huggingface.co/datasets/infgrad/retrieval_data_llm}{retrieval\_data\_llm} & Retrieval & s2p & zh & 32,551 & \href{https://huggingface.co/datasets/suolyer/webqa}{webqa} & Retrieval & s2p & zh & 4,988 \\
\href{https://huggingface.co/datasets/C-MTEB/AFQMC}{AFQMC} & STS & s2s & zh & 3,876 & \href{https://huggingface.co/datasets/C-MTEB/ATEC}{ATEC} & STS & s2s & zh & 11,387 \\
\href{https://huggingface.co/datasets/C-MTEB/BQ}{BQ} & STS & s2s & zh & 10,000 & \href{https://github.com/china-ai-law-challenge/CAIL2019/tree/master/scm}{CAIL2019-SCM} & STS & s2s & zh & 648 \\
\href{https://www.luge.ai/}{CINLID} & STS & s2s & zh & 2,883 & \href{https://github.com/IAdmireu/ChineseSTS}{ChineseSTS} & STS & s2s & zh & 2,497 \\
\href{https://huggingface.co/datasets/fenffef/cmnli}{CMNLI} & PairClassification & s2s & zh & 119,029 & \href{https://huggingface.co/datasets/shibing624/nli_zh}{nli\_zh} & PairClassification & s2s & zh & 185,787 \\
\href{https://huggingface.co/datasets/Fred666/ocnli}{OCNLI} & PairClassification & s2s & zh & 11,937 & \href{https://github.com/CLUEbenchmark/QBQTC/tree/main}{QBQTC} & STS & s2s & zh & 47,223 \\
\href{https://github.com/CLUEbenchmark/SimCLUE}{SimCLUE} & STS & s2s & zh & 290,699 & \href{https://huggingface.co/datasets/xnli}{XNLI (zh)} & PairClassification & s2s & zh & 74,252 \\
\href{https://huggingface.co/datasets/neuclir/csl}{CSL} & Classification & s2s, p2s & zh & 12,249 & \href{https://huggingface.co/datasets/SirlyDreamer/THUCNews}{THUCNews} & Classification & s2s & zh & 9,690 \\
\href{https://huggingface.co/datasets/fenffef/tnews}{TNews} & Classification & s2s & zh & 6,762 & \href{https://huggingface.co/datasets/C-MTEB/JDReview-classification}{JDReview} & Classification & s2s & zh & 1,232 \\
\href{https://huggingface.co/datasets/fenffef/iflytek}{IFlyTek} & Classification & s2s & zh & 8,221 & \href{https://huggingface.co/datasets/C-MTEB/OnlineShopping-classification}{OnlineShopping} & Classification & s2s & zh & 7,600 \\
\href{https://huggingface.co/datasets/C-MTEB/waimai-classification}{Waimai} & Classification & s2s & zh & 7,376 & \href{https://huggingface.co/datasets/CohereForAI/aya_dataset}{Aya Dataset} & Retrieval & s2p & multilingual & 26,292 \\
\href{https://huggingface.co/datasets/sentence-transformers/miracl}{MIRACL} & Retrieval & s2p & multilingual & 39,946 & \href{https://huggingface.co/datasets/castorini/mr-tydi}{Mr. TyDi} & Retrieval & s2p & multilingual & 46,997 \\
\href{https://huggingface.co/datasets/maximedb/paws-x-all}{PAWS-X} & STS & s2s & multilingual & 128,398 & \href{https://huggingface.co/datasets/mteb/amazon_reviews_multi}{AmazonReviews} & Classification & s2s & multilingual & 7,721 \\
\href{https://huggingface.co/datasets/mteb/amazon_counterfactual}{AmazonCounterfactual} & Classification & s2s & multilingual & 8,323 & \href{https://huggingface.co/datasets/mteb/multilingual-sentiment-classification}{MultilingualSentiment} & Classification & s2s & multilingual & 9,804 \\
\href{https://huggingface.co/datasets/mteb/amazon_massive_intent}{Amazon Massive Intent} & Classification & s2s & multilingual & 7,832 & \href{https://huggingface.co/datasets/mteb/amazon_massive_scenario}{AmazonMassiveScenario} & Classification & s2s & multilingual & 7,078 \\
\href{https://huggingface.co/datasets/mteb/mtop_domain}{MTOPDomain} & Classification & s2s & multilingual & 9,610 & \href{https://huggingface.co/datasets/mteb/mtop_intent}{MTOPIntent} & Classification & s2s & multilingual & 7,952 \\
\href{https://huggingface.co/datasets/wecover/OPUS_TED2020}{Bible\_nlp} & BitextMining & s2s & multilingual & 71,027 & \href{https://huggingface.co/datasets/wecover/OPUS_TED2020}{OPUS\_TED2020} & BitextMining & s2s & multilingual & 12,556 \\
\href{https://huggingface.co/datasets/aisingapore/SEA-Instruct-2602}{SEA-Instruct-2602} & InstructionRetrieval & s2p & SEA & 6,831,383 & \href{https://huggingface.co/datasets/iapp/iapp_wiki_qa_squad}{Iapp\_wiki\_qa\_squad} & Retrieval & s2p & th & 5,761 \\
\href{https://huggingface.co/datasets/taidng/UIT-ViQuAD2.0}{UIT-ViQuAD2.0} & Retrieval & s2s & vi & 28,454 & \href{https://github.com/muhammadravi251001/ac-iquad/raw/main/data/ac_iquad.zip}{Ac\_iquad\_simple\_source} & Retrieval & s2s & id & 116,640 \\
\href{https://huggingface.co/datasets/ContextSearchLM/context_search_vietnamese_prompt_97_minilmtok}{Content\_search\_vietnamese} & Retrieval & s2p & vn & 1,600,983 & \href{https://raw.githubusercontent.com/ThuraAung1601/AskCovidDrBot/refs/heads/main/data/covid_qA.csv}{Covid\_qa\_dataset} & Retrieval & s2s & my & 454 \\
\href{https://huggingface.co/datasets/muhammadravi251001/squadid-nli}{Squdid\_nli} & PairClassification & s2s & id & 236,890 & \href{https://huggingface.co/datasets/jcblaise/newsph_nli}{Newsph\_nli\_dataset} & PairClassification & s2s & fil & 281,321 \\
\href{https://raw.githubusercontent.com/ir-nlp-csui/indonli/main/data/indonli/train.jsonl}{Indonli\_dataset} & PairClassification & s2s & id & 3,476 & \href{https://huggingface.co/datasets/airesearch/wangchanx-seed-free-synthetic-instruct-thai-120k}{Wangchan120k} & InstructionRetrieval & s2p & th & 118,898 \\
\href{https://huggingface.co/datasets/airesearch/WangchanThaiInstruct}{Wangchaninstruction} & InstructionRetrieval & s2p & th & 32,207 & \href{https://huggingface.co/datasets/sarahlintang/Alpaca_indo_instruct}{Alpaca\_indo\_instruct} & InstructionRetrieval & s2p & id & 52,002 \\
\href{https://huggingface.co/datasets/NLPinas/EMoTES-3K}{Emotes\_3k} & Classification & s2s & fil & 2,905 & \href{https://github.com/vistec-AI/generated_reviews_enth/raw/main/data.zip}{Generated\_review\_th\_en} & Classification & s2s & th & 141,369 \\
Total &  &  &  &  &  &  &  &  & 14,316,233 \\
\hline
\end{tabular}}
\vspace{-3mm}
\caption{The instruction text datasets. Note that we did not update any instances in the datasets; we only added instructions for each sample to make it specific for text embeddings.}
\label{tab:instruction_datasets}
\vspace{-3mm}
\end{table*}

\end{document}